# Accelerating Drug Safety Assessment using Bidirectional-LSTM for SMILES Data

K. Venkateswara Rao[1], Dr. Kunjam Nageswara Rao[2], Dr. G. Sita Ratnam[3]

[1] Research Scholar, [2] Professor

Department of Computer Science and Systems Engineering,

Andhra University College of Engineering AUCE(A),

Visakhapatnam-530003, Andhra Pradesh, India

[3] Professor, Chaitanya Engineering College,

Madhurawada, Visakhapatnam, Andhra Pradesh-530048, India.

**Abstract**— Computational methods are useful in accelerating the pace of drug discovery. Drug discovery carries several steps such as target identification and validation, lead discovery, and lead optimisation etc., In the phase of lead optimisation, the absorption, distribution, metabolism, excretion, and toxicity properties of lead compounds are assessed. To address the issue of predicting toxicity and solubility in the lead compounds, represented in Simplified Molecular Input Line Entry System (SMILES) notation. Among the different approaches that work on SMILES data, the proposed model was built using a sequence-based approach. The proposed Bi-Directional Long Short Term Memory (BiLSTM) is a variant of Recurrent Neural Network (RNN) that processes input molecular sequences for the comprehensive examination of the structural features of molecules from both forward and backward directions. The proposed work aims to understand the sequential patterns encoded in the SMILES strings, which are then utilised for predicting the toxicity of the molecules. The proposed model on the ClinTox dataset surpasses previous approaches such as Trimnet and Pre-training Graph neural networks(GNN) by achieving a ROC accuracy of 0.96. BiLSTM outperforms the previous model on FreeSolv dataset with a low RMSE value of 1.22 in solubility prediction.

**Keywords**-  BiLSTM, SMILES, RNN, GNN, Trimnet.

## 1. Introduction

In the current landscape, bringing a new drug to market typically requires around a decade of rigorous research, development, and regulatory processes. Additionally, the cost associated with this endeavour is substantial, averaging between $2 billion to $3 billion. Drug Discovery typically begins with the identification of a biological target, such as a protein or enzyme associated with a specific disease. Scientists then search for molecules, often from natural or synthetic sources, that can interact with the target in a way that modifies its activity, leading to a therapeutic effect. Discovered molecules frequently fail to progress as potential drugs due to challenges such as toxicity, inadequate activity, and poor solubility, underscoring the complexity of drug discovery and the need for rigorous screening and optimization processes. Machine learning models are frequently employed today to predict the properties of potential drugs, offering faster results compared to manual methods. Currently, studies are utilizing a range of neural network architectures to explore the Quantitative Structure Activity Relationship (QSAR) of molecules. Artificial Neural Networks have proven to be highly efficient in analyzing QSAR based on descriptors [1]. The rapid evolution of neural network architectures has revolutionized the study of QSAR, with methodologies such as CNNs, RNNs, and BNNs offering early prediction of pharmaceutical properties of drugs such as toxicity, solubility etc.,

393

among the pharmaceutical properties' toxicity plays a critical role in the rejection of approximately one-third of drug candidates, significantly contributing to the elevated costs of drug development. The proposed model utilizes Paracelsus' principle to predict toxicity at doses relevant to patient use, distinguishing between toxic and non-toxic effects based on dosage levels for each drug [2].

SMILES (Simplified Molecular Input Line Entry Specification) is a specification in the form of line notations for describing the structure of chemical species. A chain of letters, numbers and characters that specify the atoms, their connectivity, bond order and chirality [3]. These SMILES are taken as input for the proposed model whereas in the graph model, these are converted to molecular graphs to train the model. But in the proposed model we use a tokenizer object from the TensorFlow Keras specifically designed for character-level tokenization of SMILES strings.

A machine learning (ML) algorithm capable of precisely characterizing the compositions of behavioural components can meet this requirement. By employing ML techniques, it becomes possible to assess a considerable number of materials without the need for physical samples and to efficiently ascertain their physical properties, like solubility. Machine Learning Techniques such as Random Forest, Multilinear regression and some other regression models were used previously. But the main obstacle is the final output RMSE (root mean square error) is greater than 2. By using ML approaches the error is more [4].

Recent advancements in cheminformatics have witnessed a surge in the application of deep learning techniques, leveraging computer vision, natural language processing, and other methodologies to enhance the accuracy of molecular property prediction. These approaches are categorized into two main types: sequence-based methods and graph-based methods. In sequence-based methods, such as RNNs or CNNs, molecular sequences like SMILES are effectively processed to extract meaningful representations [5]. Graph-based methods utilize techniques such as Graph Neural Networks (GNN) or High Dimensional Neural Networks (HDNN) to map molecular structures to their corresponding properties. GNNs are particularly adept at converting molecular graphs into node and edge embeddings, enabling high-performance predictions across various tasks [6-10]. In general, the proposed model is mainly used for NLP tasks. There are different NLP models like spacy, Word2Vec, and Fuzzy – Wuzzy for sequence matching or string similarity [11]. These techniques can also be used on SMILES.

## 2. About Dataset

The Clintox dataset is a valuable resource utilised for investigating the clinical toxicity of chemical compounds. It encompasses critical information about two key toxicity endpoints: clinical trial toxicity and FDA approval status. With a collection of 1491 compounds, this dataset serves as a fundamental tool for the early anticipation and assessment of toxicity during the development stages of pharmaceuticals. The proposed model is made to work on other datasets like TOX21 and synthetic data. The Tox21 dataset is characterized by its multimodal nature, encompassing diverse data from multiple sources and formats. It comprises chemical structures, molecular descriptors, and activity data stemming from 12 distinct toxicological assays conducted on 7,831 compounds. Synthetic data is made by combining both datasets tox21 and Clintox.

The FreeSolv dataset is a freely available dataset commonly used for benchmarking molecular property prediction models, particularly those related to solvation-free energies. It contains a collection of small organic molecules along with their experimental solvation-free energies in water. Each molecule in the dataset is represented by its SMILES string (a compact textual representation of a molecule's structure) and the corresponding experimental solvation-free energy in kcal/mol. FreeSolv dataset uses water as a solvent for calculating the solvation-free energies.

### 3. NEURAL NETWORKS in QSAR

QSAR, or Quantitative Structure-Activity Relationship analysis, is a crucial aspect of ligand-based screening in drug discovery. It involves understanding how the structure of molecules relates to their biological effects. Ligand-based screening focuses on the chemical features of known active compounds to predict the activity of new ones. By recognizing patterns and similarities in compound structures, these methods help forecast the activity of novel compounds.

The proposed model uses the QSAR approach to predict the toxicity and solubility of the new lead compounds by training on the known compound's data. For predicting toxicity, we can name the approach as QSTR approach which means Quantitative Structure Toxicity Relationship. In the proposed model Recurrent Neural Networks (RNN) are used. Among different RNNs such as GRU's and LSTM's, a variety of LSTM which is BiLSTM is used for the model building.

### 4. Methodology

This paper presents an analysis of diverse QSAR methodologies utilized in toxicity determination, followed by a comparison with a deep learning model based on SMILES for toxicity and solubility assessment. The current model BiLSTM represents an advanced form of the LSTM architecture, capable of processing input sequences in both forward and backward directions simultaneously, thereby enhancing its ability to find the correlation between the sequences. The input sequential data has been encoded into arrays of binary digits to facilitate processing by our model. These encoded inputs are then fed into the BiLSTM layer, which processes them bi-directionally. Finally, the output from the BiLSTM layer is passed through a dense network equipped with a sigmoid function to predict toxicity. This model aims to predict the toxicity of the compounds, and it's called Quantitative Structure – Toxicity Relationship (QSTR). It's all about understanding how the structure of molecules relates to their toxicity.

Significant progress has been made in predicting molecular properties, especially through graph-based methods. These methods start by converting SMILES inputs into molecular structures and then into molecular graphs using tools like RDKit. In these graphs, atoms act as nodes and chemical bonds serve as edges. Each node in the molecular graph is linked to a feature vector containing key atomic details like atomic number, hybridization state, and the presence of functional groups. Similarly, edges carry features representing bond type, distance, and other characteristics. The heart of graph-based models is the graph convolutional layer, which combines information from nearby nodes and edges to update node features. This iterative process improves node representations by considering information from neighboring atoms and bonds. After processing the molecular graphs through multiple graph convolutional layers, the final node representations are inputted into a fully connected neural network. These networks gather insights from all nodes in the graph and produce toxicity predictions for each molecule.

The proposed model takes a sequenced-based approach, prioritizing the examination of correlations between the sequences of molecules during toxicity prediction. Starting with the input SMILES string, SMILES provide a standardized way of representing atoms and bonds along with their arrangements within a molecule. In a SMILES representation, Atoms are denoted by their atomic symbols (e.g., C for carbon, O for oxygen). Bonds between atoms are represented by various symbols Single bonds are represented by '-'(these are not explicitly specified). Double bonds are represented by '='. Triple bonds are represented by '#'. Aromatic bonds are represented by lowercase letters (e.g., 'c' for aromatic carbon). Branching and cyclic structures are indicated using parentheses '(' and ')'. Hydrogen atoms are usually omitted and assumed to be implicitly present to satisfy valency requirements. For example, SMILES representation for ethanol (CH3CH2OH) is 'CCO'. SMILES are encoded into binary arrays and then these binary arrays with the target label are inputted into the neural network for analysis and prediction. This process allows the neural network to learn patterns and relationships between the molecular structures encoded in the SMILES and their associated properties.

*A.  Long Short-Term Memory*

It is a type of neural network that is good at learning patterns and relationships in sequences of data, like text or time series. Unlike standard feedforward neural networks, which transfer data forward after processing, LSTM (Long Short-Term Memory) networks have feedback connections. These connections allow LSTM networks to store the results of the current input for use in the near future when making other predictions. This ability to retain and selectively utilize information over time makes LSTMs particularly effective for tasks involving sequential data, such as natural language processing and time series prediction [7]. LSTM is applicable, especially for tasks like text recognition, speech recognition etc. LSTM was created to address the challenge of retaining information over longer periods, unlike other deep learning models. Its unique design allows it to remember crucial details for extended durations, making it effective for tasks where understanding sequences over time is important, like language translation or sentiment analysis. It uses a gate mechanism similar to logic gates there are three gates in main input, forget, and output gates and one more important aspect is cell state which is like a memory to LSTM. The input gate decides which information from the current input should be stored in the cell state. It controls the flow of new information into the cell. Forget gate decides which information from the previous cell state should be forgotten or discarded. It helps the model decide what to remember and what to forget from long-term memory. The output gate decides what information from the current cell state should be output to the next layer in the network. It helps the model decide what information to use for predictions. Three gates of LSTM are sigmoid activated, this activation ensures that the gate values fall within the range of 0 and 1. In practical terms, a value of 0 indicates blocking or inhibiting the flow of information, while a value of 1 signifies allowing the information to pass through the gate.

Gates equations are as follows:

$$f_t = \sigma(W_f \cdot [h_{t-1}, x_t] + b_f) \qquad (1)$$

$$i_t = \sigma(W_i \cdot [h_{t-1}, x_t] + b_i) \qquad (2)$$

$$o_t = \sigma(W_o \cdot [h_{t-1}, x_t] + b_o) \qquad (3)$$

$$\tilde{c}_t = \tanh(W_C \cdot [h_{t-1}, x_t] + b_C) \qquad (4)$$

$f_t$, $i_t$, $o_t$, $\tilde{c}_t$ is the forget gate, input gate, output gate and candidate gate output at time step t respectively, σ represents the sigmoid activation function, and W is the weight matrix for the respective gates. $h_{t-1}$ is the previous hidden state, $x_t$ is the current input, and b is the bias term of corresponding gates.

The final states are represented as:

$$C_t = f_t * c_{t-1} + i_t * \tilde{c}_t \qquad (5)$$
$$h_t = o_t * \tanh(c_t) \qquad (6)$$

Where $C_t$, represents the cell state at time t and $h_t$ is the final output of the LSTM cell. Figure 2 represents the various gates at a given time t, by giving values into the above equation's gates can be analyzed.

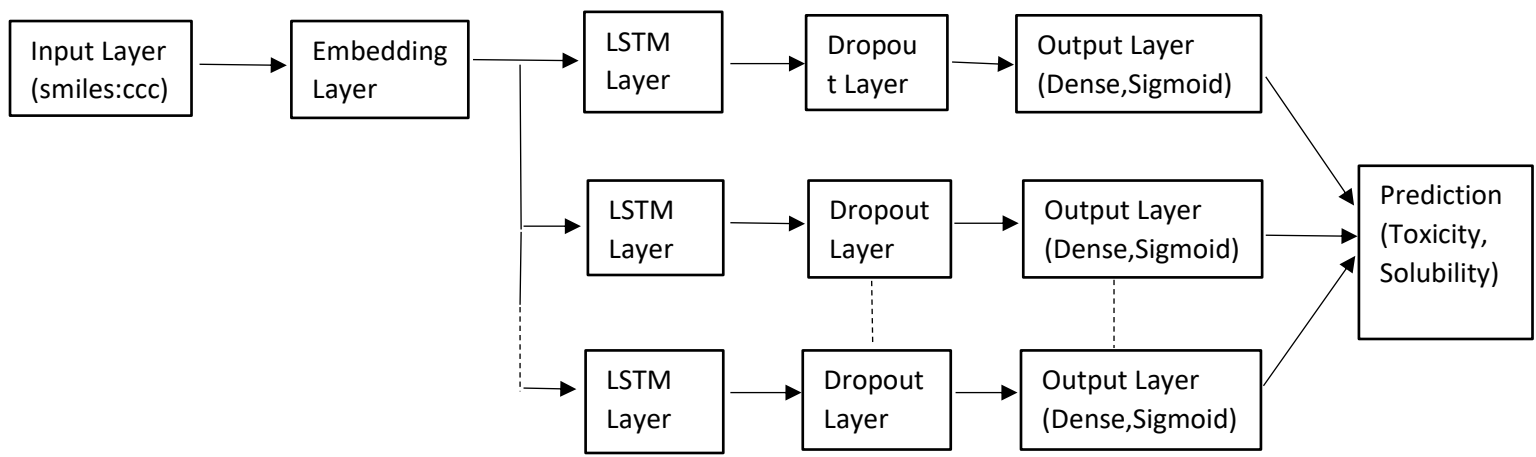

Figure 1: Architecture of proposed model using LSTM

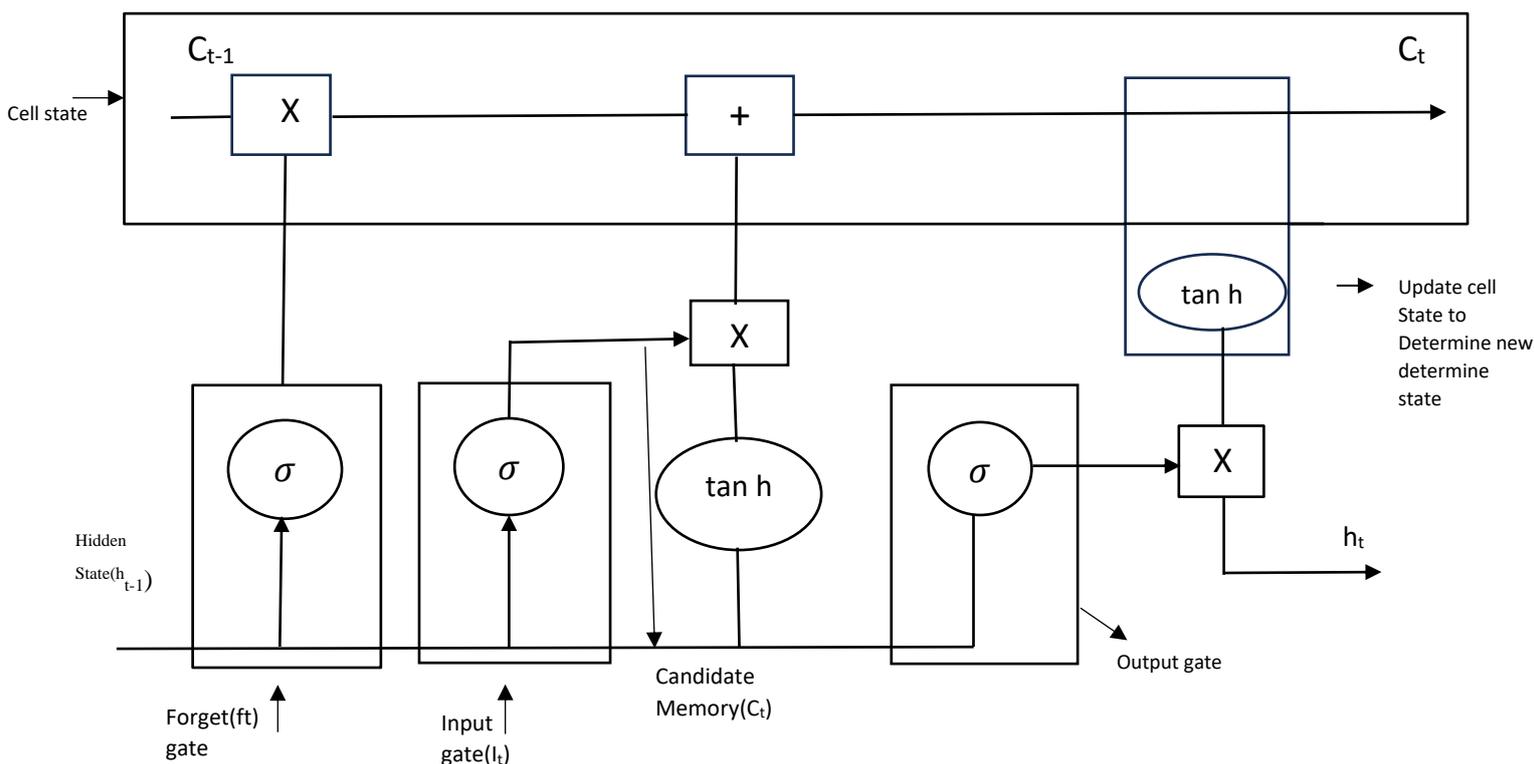

Figure 2. LSTM layer at a timestep t

From above figure 1 architecture is discussed in the form of three layers. The First layer Embedding Layer converts each SMILES sequence into a dense vector representation suitable for processing by the LSTM layer. LSTM layer processes the embedded SMILES sequences, capturing dependencies and patterns within the data over time. Finally, the Dense layer performs the final classification based on the LSTM's output, predicting the toxicity which is a binary label. Similarly, the same procedure is followed on the regressive dataset FreeSolv to find the regressive values of molecular solubility in mols per litre.

*B. Bi-directional LSTM*

From the Figure 1 architecture which is segregated into three layers, BiLSTM architecture will have more layers as it passes the information bidirectionally, it includes loss function one more step. The loss function calculates the discrepancy between the predicted outputs and the ground truth labels, providing feedback to the model on how to adjust its parameters (weights and biases) to minimize this discrepancy. The complete flow from the inputs (i.e., SMILES) taken to the model and the output i.e., prediction of toxicity label is shown in below architecture figure 2.

Bidirectional LSTMs process data in both directions simultaneously, from past to future (forward direction) and from future to past (backward direction). The first layer is the Input Layer where SMILES strings, which represent molecular structures, are fed into the network. The second layer Embedding layer where each character or token in the SMILES string is converted into a dense vector representation through an embedding layer. This dense representation captures the semantic meaning of each character in the context of the molecular structure. The next layer is Bidirectional LSTM Layer. The embedded SMILES sequences are passed into a Bidirectional LSTM layer. This layer consists of two LSTM networks, one processing the input sequence in the forward direction (from start to end) and the other processing it in the backward direction (from end to start). The Bidirectional LSTM captures both past and future dependencies in the SMILES sequences, allowing the network to understand the context of each character/token based on its surrounding characters/tokens. Finally, the Output layer is where the hidden states from both the forward and backward LSTM networks are combined to obtain the final output. This output here is predicting molecular properties (i.e., toxicity, solubility). Three gates of LSTM are sigmoid activated, This activation ensures that the gate values fall within the range of 0 and 1. In practical terms, a value of 0 indicates blocking or inhibiting the flow of information, while a value of 1 signifies allowing the information to pass through the gate.

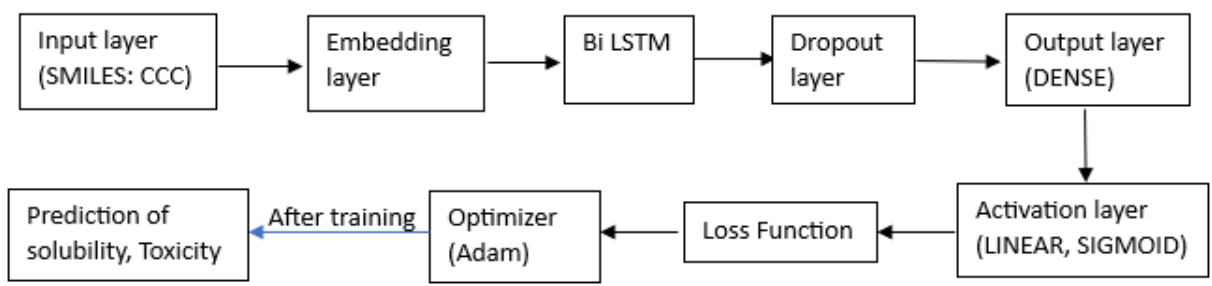

Figure 3. Architecture of proposed model using BiLSTM

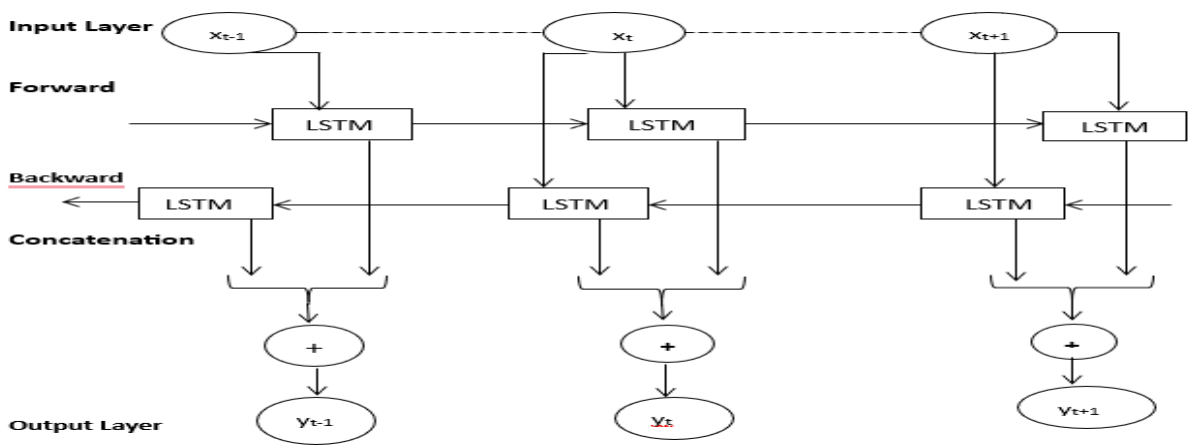

Figure 4. BiLSTM model at a timestep t

Figure 4 clearly shows the forward and backward pass of the BiLSTM and in each pass, there are many LSTM and these working is shown in figure 2.

This architecture enables the model to capture information from both past and future contexts, which can lead to better performance. During training, the parameters of the Bi-LSTM model are updated using gradient descent optimization algorithms, i.e., Adam, to minimize a loss function. Common loss functions used in sequence prediction tasks include mean squared error (MSE) for molecular solubility prediction which is a regression task.

## 5. Results

Comparative analysis demonstrates the effectiveness of the sequence-based approach in solubility prediction. Bi-LSTM models trained on SMILES data outperform traditional methods, yielding superior prediction accuracy and efficiency.

The Clintox dataset has imbalanced data, it has been balanced by using the undersampling technique. We compared models built on different methods, and our BiLSTM model, based on a sequence-based approach, achieved the highest ROC accuracy. Specifically, on the Clintox dataset, our model performed the best with an ROC accuracy of 0.96 on the FDA approval task and an average of $0.96 \pm 0.01$ ROC accuracy when considering both tasks (i.e., FDA-approved, CT-Tox), outshining other models.

Below figure 5 which contains bar graph clearly indicates the difference between the model the previous highest is the trimnet model and the proposed model achieved 2% more ROC accuracy. The proposed model is also trained on other datasets like tox21 and synthetic data.

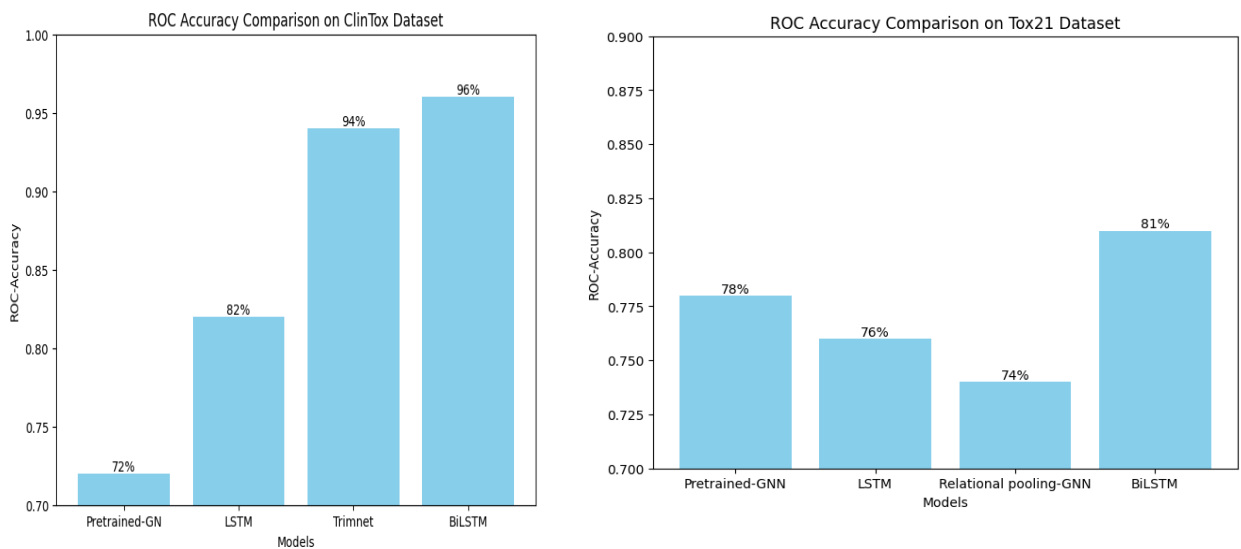

Figure 5. ROC Accuracy comparison of models on ClinTox and Tox21 datasets.

Similarly, on the Tox21 dataset, our model achieved a ROC accuracy of 0.81. surpassing graph-based methods like pretraining and relational pooling. Notably, BiLSTM outperformed GAN models in terms of ROC accuracy. When performed on synthetic data our model achieved an ROC accuracy of 0.80.

By capturing complex relationships between molecular structures and solubility, these models offer significant advancements in predictive performance. The proposed model outperforms the previous best model GLAM with RMSE difference of 0.1. where GLAM RMSE value is 1.31 [3]. and proposed model achieved 1.2 RMSE. The lower RMSE value indicates the better model. The proposed model compared with few machine learning algorithms which is shown in below figure.

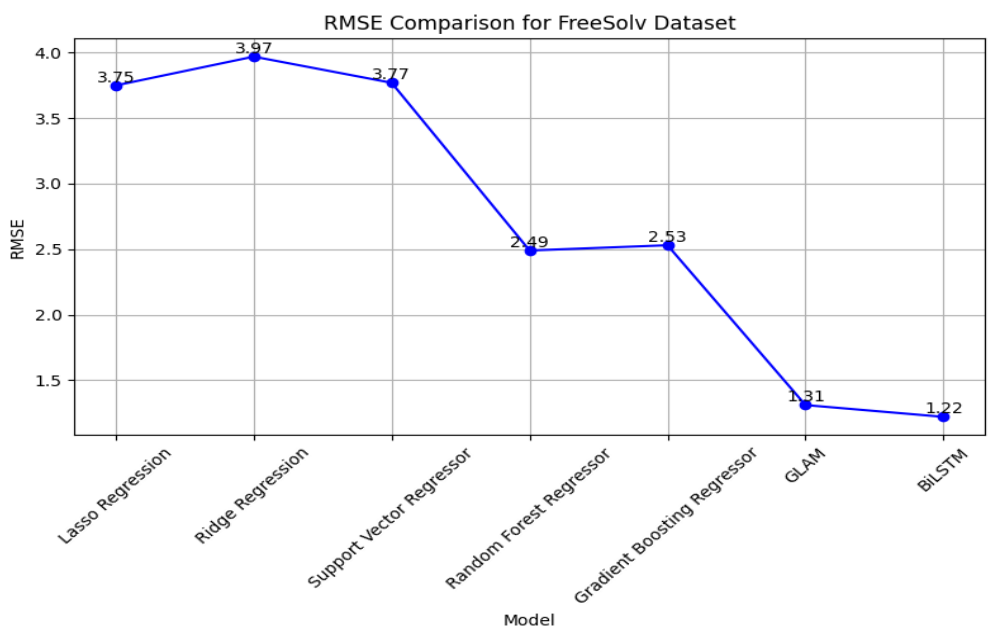

Figure 3. Comparison of BiLSTM with previous regression models.

The proposed model was taken to several experiments with various parameters to optimize our model's performance on the training data. Through these experiments, we observed significant changes in test metrics such as accuracy and Area Under the Curve of Receiver Characteristic Operator (AUROC) across different parameter configurations. After hyperparameter tuning, we found that our model achieved better roc-accuracy with the following parameters: 'dropout_rate': 0.3, 'learning_rate': 0.1, 'units': 32.

### 6. Conclusion

Sequence-based approaches, particularly Bi-LSTM networks, offer a promising avenue for enhancing solubility and toxicity prediction efficiency in drug discovery. By leveraging SMILES information, these models provide a more accurate and streamlined approach to predicting solubility compared to traditional methods. This research shows the potential of sequence-based methods in advancing computational drug discovery techniques and underscores the importance of incorporating machine learning approaches in predictive modelling tasks. The proposed model presents a computational approach for predicting drug toxicity based on SMILES representations, aiming to accelerate the drug discovery process. Deep learning methods are explored to enhance the accuracy of toxicity prediction. BiLSTM, a type of recurrent neural network (RNN), stands out for its strong performance, especially when measured using the ROC accuracy metric on the ClinTox dataset. LSTM models which are used for NLP(Natural Language Processing) tasks majorly, are used on SMILES for toxicity and solubility prediction. The sequence-based approach can be further enhanced by using models like gpts, and LLMs etc., the proposed model can be further enhanced by collecting and training on the larger amount of data.